\documentclass[runningheads,fleqn]{cl2emult}

\usepackage{makeidx}  
\usepackage{graphicx} 
\usepackage{subeqnar} 
\usepackage{multicol} 
\usepackage{engi}     

\usepackage{amsfonts}   

\newcommand{\rpred}{\mbox{\em robot-pred}}
\newcommand{\ppred}{\mbox{\em pixel-pred}}
\newcommand{\psucc}{\mbox{\em pixel-succ}}
\newcommand{\rsucc}{\mbox{\em robot-succ}}
\newcommand{\nil}{\mbox{NIL}}

\def \qed {\hfill $\Box$}

\begin{document}
\title*{Algorithms for Rapidly Dispersing Robot Swarms in Unknown Environments}
\titlerunning{Rapidly Dispersing Robot Swarms}
\author{Tien-Ruey Hsiang\inst{1}
\and Esther M. Arkin\inst{1}
\and Michael A. Bender\inst{1}
\and \protect\newline S\'andor P. Fekete\inst{2}
\and Joseph S. B. Mitchell\inst{1}}
\authorrunning{Hsiang et al.}
\institute{Stony Brook University, Stony Brook, NY 11794, USA
\and TU Braunschweig, 38106 Braunschweig, Germany}
\newcommand{\opt}{\mbox{\sc opt}}

\maketitle

\begin{abstract}
  We develop and analyze algorithms for dispersing a swarm of
  primitive robots in an unknown environment, $R$. The primary
  objective is to minimize the {\em makespan\/}, that is, the time to
  {\em fill\/} the entire region.  An environment is composed of
  {\em pixels\/} that form a connected subset of the
  integer grid.  There is at most one robot per pixel and robots move
  horizontally or vertically at unit speed. Robots enter $R$ by means
  of $k\geq 1$ {\em door pixels\/}
  Robots are primitive finite automata, only having local
  communication, local sensors, and a constant-sized memory.  

  We first give algorithms for the single-door case (i.e., $k=1$), analyzing
  the algorithms both theoretically and experimentally
  We prove that
  our algorithms have optimal makespan $2A-1$, where $A$ is the area of~$R$.

  We next give an algorithm for the multi-door case ($k\geq 1$), based on a
  wall-following version of the leader-follower strategy.  We prove
  that our strategy is $O(\log (k+1))$-competitive, and that this
  bound is tight for our strategy and other related strategies.
\end{abstract}

\section{Introduction}

{The objective of swarm robotics is to program
a huge number of
simple, tiny robots to perform complex tasks collectively.
A typical application scenario for a robot swarm may involve
two phases:}
first, the robot swarm fills an environment as quickly as
possible while keeping the swarm ``connected'' (a term made more
precise later).  Once the robots are distributed in the environment, the
robots perform {\em world-embedded\/} computational tasks,
such as computing distances and short paths
between points in the environment, finding chokepoints, mapping
the environment,
and searching for intruders.

This paper develops efficient
algorithms for filling an environment with a swarm of robots,
thus focusing on the first phase described above.  Because
target applications require thousands to millions of robots and
current demonstration scenarios have at most hundreds of robots, it is
crucial to develop algorithms
that have explicitly stated performance guarantees, thus ensuring
scalability.

There has already been work on dispersion algorithms, but most
dispersion algorithms rely on greedy strategies such as
{\em go-for-free space\/}, where robots move to fill unoccupied
space, or {\em artificial physics\/} strategies, where neighboring
robots exert ``forces'' on each other:  repulsion forces if the robots
are closer than the target separation, and attraction forces if the
neighboring robots are too far away.
Although such robot
behaviors often lead to reasonable dispersion, there are cases
in which the swarm enters infinite loops, never filling the domain.
Even in artificial physics strategies, which seem to disperse robots
most efficiently, there are no guarantees
on filling time.  Furthermore, the swarm behavior
under artificial-physics rules results in dispersions that
are more reminiscent of molecular
dynamics simulations (with large amounts of ``bouncing'' and ``jitter'')
than of the ideal coordinated behavior of cooperating teams of
robots.

The dispersion algorithms in this paper represent a departure from
previously studied dispersion strategies.  First of all, we test our
algorithms in arbitrary polygonal environments, rather than
restricting ourselves to rectangles or the
infinite plane.  It is important
to study the dispersion algorithms in complicated polygonal
domains because the target domains may be bottleneck-filled
indoor environments or systems of pipes or ductwork.

\begin{figure}
\centerline{\hfill
\includegraphics[width=.3\textwidth]{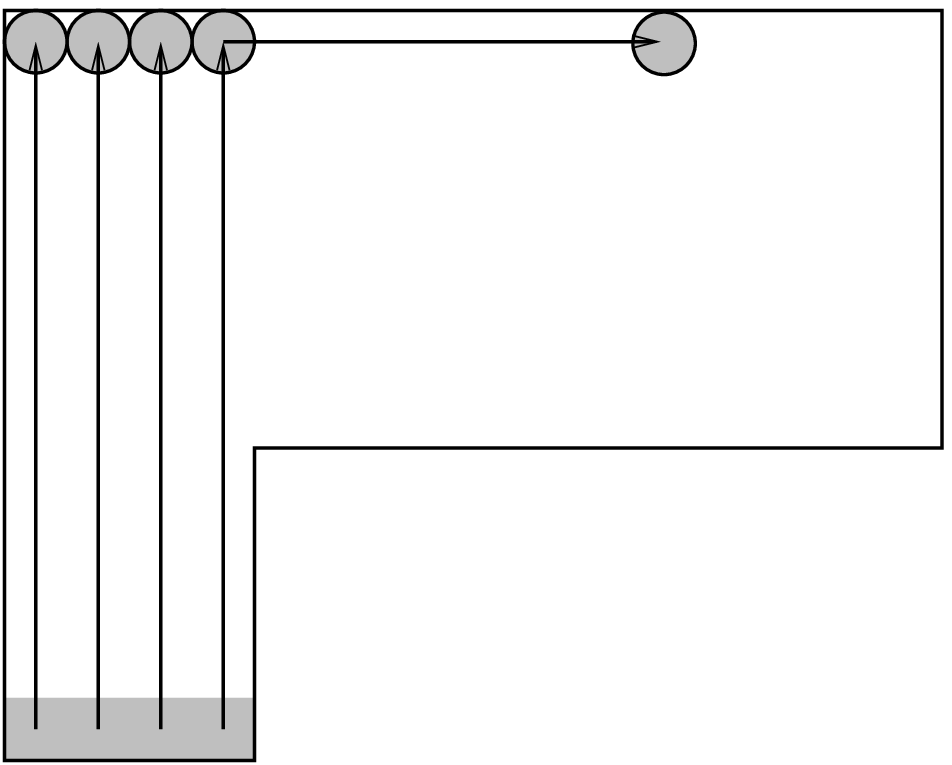}\hfill
\includegraphics[width=.4\textwidth]{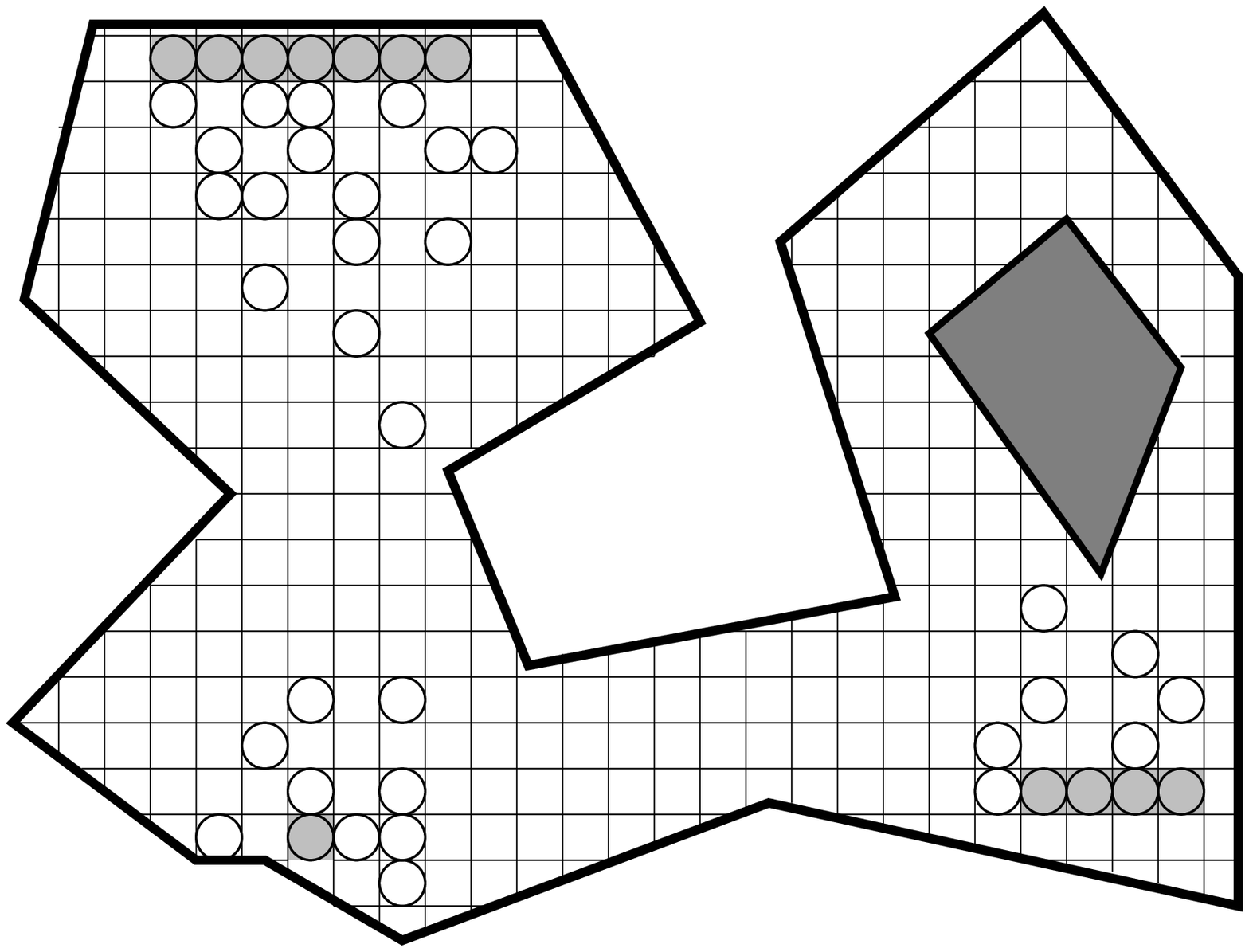}\hfill}
\label{fig:cut-off-domain-pixels}
\caption{Left: One turn can cut off most of the flow
lines leading out of the door pixels (shaded grey): The left three
columns of robots are trapped. 
Right: A polygonal domain encloses a discrete set of possible
robot locations ({\em pixels}); special door pixels are shown in grey.}
\end{figure}

Our algorithms are based on follow-the-leader local rules, where
the robots form chains of leaders and followers emanating from
a central source of robots.
(the ``doors'').
The emergent behavior of
our main algorithm is a left-hand-on-the-wall strategy, where the entire
robot swarm hugs the left wall of the domain as the filling proceeds.
We provide matching lower and upper bounds on the amount of time to
fill as a function of the total size of the source.
The difficulty in
dispersing efficiently is in sustaining a large flow
of robots coming in through the doors; the ``chains'' of robots coming
in through the doors have a tendency to ``cut each other off''; see
Figure~\ref{fig:cut-off-domain-pixels}(left).  Thus, we devise
a set of local
rules for splitting and splicing the robot chains. Our experience with
our robot simulator shows that
unless the local rules are precisely
implemented, the flow of robots out of the source diminishes and
perhaps even grinds to a halt prematurely.

\subsection{Our Results}
We develop and analyze algorithms for dispersing robot
swarms in an unknown environment $R$. The primary
objective is to minimize the {\em makespan\/}, that is, the time to {\em
fill\/} the entire region.  An environment is composed of {\em
squares\/} or {\em pixels\/} that form a connected subset of the integer
grid; such an environment is also called a {\em maze\/}.
There is at most one robot per pixel and robots move
horizontally or vertically at unit speed. There is a source of robots
entering the environment through $k\geq 1$ {\em doors\/}.
Robots are primitive finite automata, only having local
communication, local sensors, and a constant-sized memory.
The challenge
is in proving that purely local
strategies distributed over a swarm of agents can result in a
predictable and provable behavior for the collective.

This paper presents the following results:
\begin{itemize}
\item We give filling algorithms for the single-door case (i.e., $k=1$),
  analyzing the algorithms both theoretically and experimentally in terms of
  natural performance metrics, such as makespan and
  total traveled distance by all robots.  Our algorithms are
  based on leader-follower strategies that are adapted from
 depth-first and breadth-first search
 to apply to  robot swarms.  We prove that our algorithms have
  optimal makespan $2A-1$, where $A$ is the area (number of
  pixels) of~$R$.

\item We give a filling algorithm for the multi-door case ($k\geq 1$),
  also based on a leader-follower strategy.
The emergent behavior of the
swarm is a filling strategy, called
{\em laminar flow\/},
 that
generalizes the ``left-hand-on-wall'' strategy to multiple
streams of robots coming in through multiple doors.
An important contribution of the algorithm is the formulation of
splitting and
splicing strategies that maintain a large
flow of robots out of the doors.
We prove that the laminar flow algorithm
is $O(\log (k+1))$-competitive, 
that is,
the ratio of the makespan of our algorithm
to the optimal makespan is $O(\log (k+1))$. 

\item We give a matching lower bound of $\Omega(\log (k+1))$
on the competitiveness of the natural class of ``simple-minded''
strategies that use only strictly local information.

\item We report on experiments with a Java simulator that we built for
  swarms in grid environments. All of the algorithms in this paper are
  implemented in this simulator.
\end{itemize}

While our results here are given purely in terms of robots moving
synchronously on discrete grids, the results apply also to dispersing
swarms of robots within an arbitrary connected planar domain, in the
ideal setting in which the robots have perfect motion control and
synchrony.  More importantly, our results may form the foundation for
theoretical work on more realistic models of swarm robotics in the
continuum, with asynchronous motion, sensor errors, slippage, and
hardware failures.

\subsection{Prior and Related Work}

There has been considerable study recently of distributed control and
coordination of a set of autonomous robots.  
Gage~\cite{dwgage-website,gage92sensor} 
has proposed the development of command and control tools for
arbitrarily large swarms of microrobots and has proposed coverage paradigms
in the context of robot dispersal in an environment.  Payton et
al~\cite{payton-est-how2002,payton-daily-est-how2001} propose the
notion of ``pheromone robotics'' for world-embedded computation.
Wagner et
al.~\cite{bruckstein97probabilistic,wagner95cooperative,wagner96cooperative,wagner98efficiently}
developed multi-robot algorithms, directly inspired by ant behaviors,
for searching and covering.
Principe et al~\cite{fpsw-dcsam-00,p-dcsam-02} and Suzuki et
al~\cite{DumitrescuSuzukiYamashita2002,sugihara96distributed,suzuki99distributed} have studied algorithmic aspects of
pattern formation in distributed autonomous robotics under various
models of robots with minimal capabilities.
The related flocking problem, which requires that a set of robots
follow a leader while maintaining a formation, has been studied
in several recent papers; see, e.g., \cite{ba-bbfcm-98,bla-bfbrb-02,gp-fsamr-01}.
Balch~\cite{teambots} has developed ``Teambots'', a Java-based
general-purpose multi-robot simulator.

There is a vast literature on algorithms for one or several robots to
explore unknown environments.  The environments can be modeled as
graphs (directed or undirected), mazes, or geometric domains, and the
robots can have a range of computing powers; see, e.g.,
\cite{AwerbuchBeRiSi95,BenderFernandezRonSahaiVadhan02,BenderSl94,DengPa90,DudekJeMiWi97}.
Mitchell~\cite{mitchell98geometric} includes a survey of many results
on exploring and navigating in geometric environments.

Dispersion algorithms have been recently studied both in the context
of multi-robot coverage and sensor network deployment; see
\cite{bs-mrdcp-02,bs-solam-02,hms-msndu-02,w-dsdcv-00}.
Methods include potential fields~\cite{hms-msndu-02,reif99social},
``artifical physics''~\cite{spears99using},
attraction-repulsion~\cite{icra03}, and algorithms based on molecular
dynamics~\cite{bs-solam-02}.  While these studies have examined
empirically the merits of heuristics, they have not addressed formally
the algorithmic efficiency of the dispersion problem.

\section{Preliminaries}
\label{sec:preliminaries}

For an arbitrary connected region ${\cal R}$ in the plane,
we let $R$ denote the set of {\em pixels\/} that lie entirely
within ${\cal R}$.  A {\em pixel\/} is an axis-aligned unit square,
$\{(x,y): i\leq x\leq i+1, j\leq y\leq j+1\}$, whose
lower left corner is given by the integer grid point $(i,j)$.
Two pixels are {\em neighbors\/} if they share a common edge;
thus, pixel $(i,j)$ has four neighbors, $(i-1,j)$, $(i+1,j)$,
$(i,j-1)$, and $(i,j+1)$.
We refer to $R$ as the {\em environment\/}, and we assume that
it is connected. A subset, $D\subset R$, of the pixels
are {\em doors\/}, which serve as sources of incoming robots.
See Figure~\ref{fig:cut-off-domain-pixels}(right).

A {\em robot\/} $r$ is said to {\em occupy\/}
a pixel $p=(i,j)$ when it is located in that pixel.
We assume that at most one robot can occupy any one pixel
at any given time $t$.  We let $p(r,t)$ denote the pixel
occupied by robot $r$ at time $t$.  We let $prev(r,t)=p(r,t-1)$
denote the pixel {\em previously\/} occupied by robot~$r$.

Time is discretized into steps $t=0,1,2,3,\ldots$.  The positions of
the robots at time $t$ are given by the set $S(t)$,
where $D \subseteq S(t)\subseteq R$.
At time $0$, there is one robot in each door: $S(0)=D$.

Each pixel is in one of three states at time $t$:
(1) the pixel is an {\em obstacle\/} if it does not lie in $R$;
(2) the pixel is {\em occupied\/} if it is in $R$ and a robot occupies it;
or (3) the pixel is {\em unoccupied} if it is in $R$ and no robot occupies it
at time $t$.
An unoccupied pixel at time $t$ is classified as either
{\em previously occupied\/}, if it was occupied by some robot
at some time prior to $t$,
or {\em frontier\/}, if it has never been occupied.

Robots have {\em sensors\/} that detect information
about the local environment.
In particular, if a robot occupies pixel $(i,j)$ at time
$t$, then we assume that it can detect the state of each pixel
within a distance $r_S$ of $(i,j)$, where $r_S$ is the (fixed)
{\em sensor radius\/}.  Robots have no global sensor (e.g., a GPS)
and no knowledge of the coordinates of the pixels they occupy.

We assume that each robot has a small finite memory
that allows it to remember the sensor readings of
the last $T\geq 1$ time steps; i.e., a robot knows the sensor
readings it has taken at time $t-T$, $t-T+1,\ldots,t-1,t$.
In particular, each robot knows which
nearby pixels have been occupied recently by other robots.
The algorithms in this paper have an $O(1)$-bit memory;
thus, the robots have the computational power of finite automata.

We assume that robots have a limited ability to communicate
with nearby robots.  In particular,
at any given time step
a robot is able to exchange a constant-size message with any robot
that lies within a prescribed {\em communication radius\/}, $r_C$,
of the robot.
 The {\em communication graph\/}, ${\cal G}(t)$,
of the swarm $S(t)$ at time $t$ is defined to be the undirected graph
whose nodes are the robots $S(t)$ and whose edges link pairs of
robots that communicate.  The swarm is said to be {\em connected\/} at time $t$ if
each connected component of ${\cal G}(t)$ contains at least one door pixel.

Robots move discretely on the grid of pixels.  If at time $t$ a robot
occupies pixel $(i,j)$ of $R$, and a neighboring pixel (say,
$(i+1,j)\,$) of $R$ is unoccupied at time $t$,
then the robot may take a step to the
right, so that at time $t+1$ it occupies pixel $(i+1,j)$.
Naturally, no two robots can move into the same pixel, since no pixel
can be occupied by more than one robot.  Thus, if two robots are
occupying pixels that neighbor an empty pixel $p$ of $R$, then the
robots must {\em negotiate\/} which one of them will have the right of
way to enter $p$.  A simple way to handle this negotiation is to
establish a priority among the four directions, e.g., north, south,
east, west.  The robot occupying the pixel with the highest priority
with respect to $p$ has first rights for moving to $p$.

The above rules on occupying and vacating pixels
force a constant {\em delay\/} in the motion.
Suppose that robot $r$ occupies
pixel $(i,j)$ at time $t$ and vacates  $(i,j)$
at time $t+1$.
No other  robot $r'$ can occupy pixel $(i,j)$
during time $t+1$; the earliest possible time that the
pixel can be occupied is $t+2$.
In principal, we could vary these local rules.
For example, we could allow robot
$r'$ to enter pixel  $(i,j)$ at timestep $t+1$,
if it enters in one direction while the current robot $r $
leaves in the opposite direction.
However, these rules would require a much higher degree of coordination
among the robots,
and the movement decisions of robots would have to be less local.
Specifically, whether a given robot
can enter one pixel may depend on whether a different robot
far away leaves another pixel.
Alternatively, we could build a longer delay into the movement rules,
but the resulting filling algorithms would be essentially the same.
Note that the delay has an immediate impact on the necessary
sensor range: If we require a delay of two time steps, robots
need at least a sensor range of two, in order to be able to
keep track of predecessors and successors.

A similar issue arises in the way new robots enter through the
door. For simplicity, we assume
that a door pixel is always occupied by a robot; as soon as the robot
occupying a door pixel moves to an adjacent pixel, a new
robot materializes at the door.
One can consider there to
be an infinite supply of robots on the ``other side'' of a door
pixel, so that a new robot steps into the door whenever the robot
that was there previously moves to another pixel of $R$.
When a robot first appears at the door, we assume that the robot points
north.

We say that the robots {\em have filled\/} the region $R$ at time $t$
if $S(t)=R$; in this case, the robots are {\em well dispersed\/}.
A {\em strategy\/} is a set of rules by which robots move, basing
their movement decisions solely on the constant amount of information they
sense and remember.  The {\em makespan\/} of a strategy
is the minimum time, $t^*$, until the robots have filled the region~$R$.

\section{Dispersion Strategies for a Single Door}

We begin with the case of a single door pixel $s$, which we call
the {\em source}.
We describe strategies based on the notion
of ``leaders'' and ``followers''.

A robot $r$ at time $t$ at position $p(r,t)$ is classified as
{\em moving\/} (meaning that it is  ``active'') or {\em stopped\/}
(meaning that it no longer moves).
If $r$ is moving, it is classified as either a {\em leader\/} or a {\em follower\/}.

In our leader-follower strategies, each robot $r$ at time $t$ has a
successor robot, $succ(r,t)$, that is following $r$ and a predecessor
robot, $pred(r,t)$, that is leading $r$.  If robot $r$ is at the door
at time $t$ ($p(r,t)=s$), then we define $succ(r,t)$ to be NIL; if $r$
is a leader, then we define $pred(r,t)$ to be NIL.

At time $t=0$ there is a robot at the door $s$, which is designated
as moving and as a leader.

\subsection{Depth-First Leader-Follower}

The depth-first leader-follower strategy (DFLF) is
inspired by depth-first search in a graph.
Specifically, at any given time $t\in\{0,1,2,3,\ldots\}$, there is
exactly one leader, $r$, which is on pixel $p(r,t)$ and $r$ is looking
for a frontier pixel (one that has never been occupied by a robot).
(Thus, necessarily the pixel from which the robot came, $prev(r,t)$,
is not frontier at time $t$.)  If there are two or more frontier
pixels neighboring $p(r,t)$, the leader selects any one of them as its
next destination.

If the leader has no frontier pixel next to it, it stops (its state
changes to ``stopped''), and it tells its successor, $succ(r)$, to
assume the leadership ($succ(r)$'s state changes to ``leader'') at the
conclusion of this time step. (The predecessor completes its move
before taking over the leadership.)  If the leader has no predecessor
(i.e., it is a robot at the door), then the algorithm halts.
(In this section, successors and (non-Nil) predecessors
are invariant over time,  and thus $t$ is removed from the argument.)

Any robot $r$ that is a follower (i.e., not the leader) simply follows
its predecessor: at time $t$ it steps next to $prev(pred(r),t)$, the
pixel previously occupied by the predecessor of $r$.
Note that
once a robot stops, it never moves again.
Furthermore, at any point in time
there is exactly one leader.

When dealing with finite automata, we have to use particular
care when implementing the leader-follower strategy.  As each robot
has only a finite number of states, it cannot keep track of the
identities of other robots, nor carry its own ``ID label''.
Instead, we use spatial information
to pursue the immediate predecessor.  Each robot $r$ has a ``heading'',
indicating the direction in which it is moving.
The predecessor $pred(r)$ is the nearest robot in that direction. Similarly,
each robot keeps track of its ``tailing'', the direction
from which it came. Thus, the successor is the nearest visible
robot in that direction. Whenever a robot is about to change its heading,
it signals this turn to its immediate successor. Heading and tailing
are updated according to delay and sensor-range. This is reflected in the
following lemma and
corollary, which are proved by introduction on time.

\begin{lemma}
If $r$ is not the leader at time $t$, then $prev(pred(r),t)$ is an
unoccupied pixel neighboring $p(r,t)$.  Furthermore, the maximum
($L_1$) distance between $r$ and $pred(r)$ is 2.
\end{lemma}

\begin{corollary}
If a pixel is not occupied
in two consecutive time steps $t$ and $t+1$, then it was
never occupied before time $t$.
(i.e., there are no ``large gaps'' of time in the occupation of a pixel.)
\end{corollary}

A further consequence of the above lemma and corollary
is that the DFLF strategy
can be implemented with a sensor radius of $r_S= 2$ and a memory of
$T=1$ (i.e., each robot recalls its previous sensor reading).
Now it is not hard to derive the following result:

\begin{theorem}
\label{th:dflf}
The DFLF algorithm halts when all pixels are occupied by robots.
The makespan of the DFLF algorithm is $2A-1$, where $A$ is the number of
pixels of the environment~$R$.
\end{theorem}

        \begin{proof}
        At each time step there is exactly one leader and that leader will do
        one of two things: (1) move to a frontier pixel, or (2) change its
        status to stopped and transfer leadership to its predecessor.  (If the leader
        has no predecessor, that is, it is at the door, the algorithm halts.)
        Since each non-door pixel can be changed from frontier to non-frontier
        at most once, and each pixel can have a robot ``park'' itself on top
        of it at most once, the number of steps before the algorithm halts is
        at most $2A-1$.  The algorithm does not stop earlier since there is
        always a leader and the leader must do one of the two actions ((1) or
        (2)) above.\qed
        \end{proof}

Note that the factor 2 is a result of using a delay of 2.
It is easy to see that this is best possible.

The total distance traveled by all robots in a filling using DFLF is
at most $A^2$, since each robot steps at most $A$.  We note that there
are examples in which {\em any\/} dispersion strategy will require
$\Omega(A^2)$ total travel.  One trivial example is a $1\times A$
rectangle, with the door pixel at one end.  A trivial lower bound on
the total travel for any given instance is the sum of the distances
from each pixel of $R$ to the door.

DFLF may use substantially more total travel than an optimal
(minimum makespan) strategy that minimizes total travel; consider, for
example, a square with side lengths $A^{1/2}$, for which DFLF requires
total travel of $A^2$ while an optimal strategy achieves total travel
$O(A^{3/2})$.

\subsection{Breadth-First Leader-Follower}

We now describe an alternative dispersion algorithm, 
the breadth-first leader-follower (BFLF)
strategy.  BFLF often has advantages over the DFLF strategy in terms
of other metrics of performance (total travel of the
robots, maximum travel of any one robot, total relative distance,
etc), while still being optimal in terms of makespan ($2A-1$).

The BFLF strategy is substantially more complex than the DFLF,
as it is not trivial to implement breadth-first search with local
decision rules on (finite-automata)
robots; indeed, our strategy does not perform
breadth-first search, but it ``approximates'' breadth-first
search sufficiently well to have some similar properties.

As before, a robot can be in a ``moving'' state or a ``stopped''
state; however, we now introduce a third state, the {\em waiting\/}
state, which is an interim state when a robot pauses
temporarily and waits to be able to move.  Once a robot is in a
stopped state, it never moves again.
In the BFLF strategy we can have multiple
leaders, while in DFLF we have only one. As described for DFLF,
we use a limited amount of spatial information to keep track
of successors and predecessors. Other local rules are more complex
and are described in the following.

Initially, there is one leader robot --- the robot at the door, $s$.
When a robot $r$ materializes at the door, it chooses to follow the
previous robot that left the door.  A leader always attempts to go to
a neighboring frontier pixel, but makes sure it does not stray too far
from its follower.  If there are no neighboring frontier pixels, the
leader waits for the immediate follower to catch up. As soon as the
immediate follower catches up, the leader becomes stopped and passes
on the leadership role to the immediate follower.  If the leader $r$
at pixel $u$ has a choice among neighboring frontier pixels, it picks
any one of them to be its heading.
If other frontier pixels adjacent to $u$ remain unclaimed by
other robots following different branches of dispersion, then the
follower $r'$ of $r$ will choose one of these pixels as its heading
when it arrives at $u$; if there remains a frontier pixel adjacent to
$u$, then $r'$'s follower $r''$ chooses this pixel as its heading when
it arrives at $u$.  Here, $r'$ and $r''$ are relabeled ``leaders'' and
pixel $u$ is marked as a branching point.

A robot $r$ is only allowed to move at time $t$, if it satisfies the
{\em Movement Criterion\/} (MC): A robot currently occupies
$prev(r,t)$ (its ``parent'' position) or a robot is currently
{\em heading for\/} $prev(r,t)$.  If the MC is satisfied, the robot
$r$ moves to its heading; otherwise, it remains at its current
position, $p(r,t)$ but is still {\em heading for\/} its target pixel.
As we will see below, a robot $r$ is never blocked by its immediate
predecessor {\em except\/} for the time step at which $r$ first
appears at the door or a time when its immediate predecessor is a
leader with nowhere to go, meaning that the leadership will be passed
to~$r$.

The BFLF strategy tries to create as many paths as possible at all times.
The visited pixels form a {\em tree\/} that guides the directions that
robots move; thus, pixels have
{\em parents\/} and {\em grandparents\/}.
At branches in the tree (pixels with $\geq 2$ children),
robots alternate the direction that they travel.
Specifically, when a robot $r$ leaves a pixel
at time $t$, it tells
its immediate follower
$r'= succ (r,t) $
what  $r'$'s
immediate destination should be.
Branching enables
robots arriving at the
same pixel at different times to go in different directions, thereby
balancing the flow.

\begin{lemma}
If not all pixels are occupied, then some robot
can move.
\end{lemma}

{The following structural lemma shows that the BFLF algorithm is
  nonblocking, i.e., a robot $r$ is never ``delayed'' by its
  predecessors, only by its followers; i.e., only the MC
  holds a robot back, not temporary blockages downstream.
  Correctness follows by induction on the height of the tree.}

\begin{lemma} 
  If a robot is at pixel $v$ and a robot is at pixel $u$, the non-root
  parent of $v$, then the robot at $v$ is stopped.
\label{clm:stopped-condition}
\end{lemma}

\begin{corollary}
A robot moves from the root every other
time step, until all pixels are occupied.
I.e., if a robot did not pop up at the root  at time $t$,
then one must pop up at time $t+1$, unless all pixels (including
the root) are occupied.
\end{corollary}

\begin{lemma}
  The robots maintain a communication distance of $3$, meaning that
  the algorithm works for $r_S\geq 3$.  The maximum $L_1$-distance
  from a robot to a follower is $3$.  Furthermore, at least one of the
  following pixels is occupied when a robot is at pixel $u$: (1) the
  parent of $u$, (2) the grandparent of $u$, or (3) the uncle of~$u$.
\end{lemma}

Putting all the claims together and using similar reasoning as for
Theorem~\ref{th:dflf}, we get the same kind of bound on the makespan:

\begin{theorem}
\label{th:bflf}
The BFLF algorithm halts when all pixels are occupied by robots.
The makespan of the BFLF algorithm is $2A-1$, where $A$ is the number of
pixels of the environment~$R$.
\end{theorem}

\subsection{Experimental Comparison}

While the DFLF and BFLF strategies both have optimal makespan, they
differ with respect to other performance metrics.
We have implemented in Java a simulator for
testing our dispersion strategies, while measuring various performance
measures, including: (1) the {\em depth} of the tree of all
  paths from the source (i.e., the length of the longest path of a
  robot); (2) the average distance traveled by a robot during the
  dispersion; and (3) the average of the ratios, $\rho_i=l_i/d_i$, of
  the ($L_1$) distance $l_i$ from the door to robot $i$'s final
  location to the total distance $d_i$ traveled by robot $i$ during
  the dispersion. 

We have compared the DFLF and BFLF strategies for
the case of a single door, in a variety of different environments.
Our results show that the average depth of the tree grows
  substantially more rapidly, as a function of the number of pixels,
  for DFLF than it does for BFLF; e.g., for an $n$-by-$n$ square grid,
  BFLF tree depth grows linearly in $n$ while DFLF depth grows close
  to quadratically in $n$.  Similarly, the average distance traveled
  by a robot using DFLF grows more rapidly than using BFLF.  The
  average of the ratios $\rho_i=l_i/d_i$ decreases with increasing
  problem size for DFLF, while it increases for BFLF.  Details of the
  experimental results are deferred to the full paper.

\section{Dispersion with Multiple Doors}

We now consider the case in which there are $k \geq 1 $ door pixels.
In order to achieve rapid filling, the objective is to maintain a flow
of robots out of as many doors as possible.  The difficulty is that
one robot chain out of one door pixel may unnecessarily block the flow
of robots out of other door pixels.  (See the introduction and
Figure~\ref{fig:cut-off-domain-pixels}(left) for examples.)

\subsection{Laminar Flow Leader-Follower (LFLF) Algorithm}
\label{sec:lflf}

This section describes the {\em Laminar Flow Leader-Follower (LFLF)\/}
algorithm.  The LFLF algorithm maintains rapid flows by careful use of
cutting and splicing.  The emergent behavior of the swarm is a
left-hand-on-wall strategy.  The behavior of the algorithm is complex
because bottlenecks in the environment may force the flow to divide
into smaller flows that fill different regions and may merge and split.

In the LFLF algorithm there are $k $ multiple door pixels,
$s_1,s_2,\cdots , s_k $.
Note that LFLF matches DFLF algorithm if $k=1$.
As in the $k = 1 $ case, our dispersion
strategies are based on leaders and followers.  A robot $r $ at time
$t$ at position $p (r, t) $ is classified as {\em active\/} or {\em
  inactive\/} (stopped).  If $r $ is active, it is either a {\em
  leader\/} or a {\em follower\/}.  However, unlike for $k = 1$,
robots {\em revert} between leaders and followers.  (In contrast, for
$k = 1 $ a robot changes roles from follower to leader but only
relinquishes the leader role by stopping and becoming inactive.)
Furthermore, in DFLF once a robot stops, it becomes inactive.  In
contrast, in LFLF a robot may temporarily stop while still remaining
active.

In LFLF (as in DFLF),
robots have successors and predecessors, but in
LFLF a robot's immediate successor or predecessor
may change over time.
Thus, each robot $r $ at time $t$
has a predecessor robot denoted
$\rpred (r, t) $
and a successor robot denoted
$\rsucc (r, t) $.
If robot $r$ is at a door pixel at time $t$
 ($p(r,t)=s_i$, for
$i\in [1,k]$), then $\rsucc (r)= \nil $;
if $r$ is a leader at time $t$, then $\rpred (r)=\nil $.

Here we further refine the roles of the pixels.
A pixel can be:
(1)~an {\em obstacle\/},
(2)~a {\em frontier\/} pixel (a pixel that has never been occupied),
(3)~an {\em inactive\/} pixel,
that is, an (occupied) pixel that hosts an inactive robot,
(4)~an {\em active\/} pixel,
(which may be occupied or not).
If a pixel is unoccupied but previously occupied,
then it is always active;
if a pixel is occupied, then it is active if and only if
the robot it hosts is active.

Note that door pixels may be active or inactive,
as described later.
Active pixels play the role of transporting robots throughout the domain
whereas inactive pixels are essentially obstacles.
Analogous to the definition of predecessor and successor robots,
each active pixel has predecessor and successor pixels.
In particular, pixel $u  $ at time step $t $
has successor pixel
$\psucc (u, t)$
and predecessor pixel
$\ppred (u, t)$.
For door pixels $s_i $ at any time $t\geq 0$
$\psucc (u, t) = \nil $.

A {\em leader pixel\/} $u $
at time $t $ has
$\ppred (u, t) =\nil $.
As we will see, if a leader pixel contains a robot,
then the robot is a leader, however the leader robot
may not be on a leader pixel and the leader pixel may not contain robots.
We call a chain of predecessor
pixels starting from the door pixel $s_i $
and ending at a leader pixel, a
{\em flow line\/}.

We now define formally the {\em left side\/} and {\em right side\/} of the flow.
To do so, we first present some intermediate definitions.
Consider a pixel $v $ at time $t$,
and let
$u =\psucc (v, t) $
and
$w =\ppred (v, t) $.
Let the {\em incoming direction\/} for $v $ at time $t$ be the vector
$\overrightarrow{v\, u}$
and let the {\em outgoing direction\/} for $v $ at time $t$ be the vector
$\overrightarrow{v\, w}$.

Consider the (vertical or horizontal) neighboring pixels, when we
start from the incoming direction $\overrightarrow{v\, u}$ and rotate
{\em clockwise\/} until reaching the outgoing direction
$\overrightarrow{v\, w}$.  Those neighboring pixels at intermediate
(vertical or horizontal) orientations are defined to be on the {\em
  left-hand side\/} of $v $ time $t$.  Thus, the left-hand side may
contain $0 $, $1 $, or $2 $ pixels.  Any neighboring pixel that is
{\em not\/} on the left-hand side, is on the {\em right-hand side\/};
this includes the pixels at the incoming and outgoing directions.

Note that leader pixels do not have outgoing edges, invalidating this
definition.  For a leader pixel, we define all neighboring pixels
(except the successor pixel) to be on the left-hand side.

The {\em left side of a flow line\/} is the union of the left sides of
the non-leader pixels in the flow line.

The essential idea the LFLF algorithm is to treat the left side of the
flow line completely differently from the right side of the flow line.
Specifically, we splice into the left-hand side but we cannot splice
into the right.

\subsection{Splicing and the LFLF Algorithm}

Suppose that at time $t$ a leader robot $r_a$ reaches a pixel $ u =
p(r_a, t)$.  If $u$ has an (unoccupied) predecessor pixel
$\ppred(u,t)$, then robot $r_a$ moves to this pixel.  (Since $r_a$ is
a leader robot, it will never have an occupied predecessor pixel.)

Otherwise, pixel $ u = p(r_a, t)$ is a leader pixel.  If $ u = p(r_a,
t)$ has no neighboring frontier pixels, then robot $r_a$ tries to pass
on the leadership.  Robot $r_a$ first looks for a neighboring active
pixel $v$.  If no such $v$ exists, then $r_a$ becomes inactive and
passes the leadership to its successor robot $\rsucc(r_a,t)$.  When
$r_a$ becomes inactive, pixel $u$ also becomes inactive and its
successor pixel becomes the new leader pixel.

If robot $r_a$ looks for a neighboring active pixel $v$, and such a
$v$ exists, then $r_a$ checks whether $ u = p(r_a, t)$ is a left-hand
neighbor of $v$.  If this is not the case for any possible $v$, then
again $r_a$ becomes inactive and passes the leadership to its
successor robot $\rsucc(r_a,t)$.  As before, when $r_a$ becomes
inactive, pixel $u$ also becomes inactive and its successor pixel
becomes the new leader pixel.

If there exists such a $v$, robot $r_a$ chooses the first such $v$
sweeping {\em clockwise\/}, starting from the incoming direction.  We
redefine the predecessor pixel $\ppred(u,t+1) := v$ and thus successor
pixel $\psucc(v,t+1) := u$.  Thus, the previous successor pixel of $v$
becomes a new leader pixel.  (Note that this new leader pixel may or
may not have a robot on it.)  Now $r_a$ identifies its new leader; it
follows predecessor pixels starting at $v$ until it finds a robot
$r_b$ and sets $\rpred(r_a,t+1)=r_b$ and $\rsucc(r_b,t+1)=r_a$. Robot
$r_a$ then passes the leadership to $r_c=\rsucc(r_b,t)$.

\begin{figure}[h!bt]
\centerline{
\includegraphics[scale=.20]{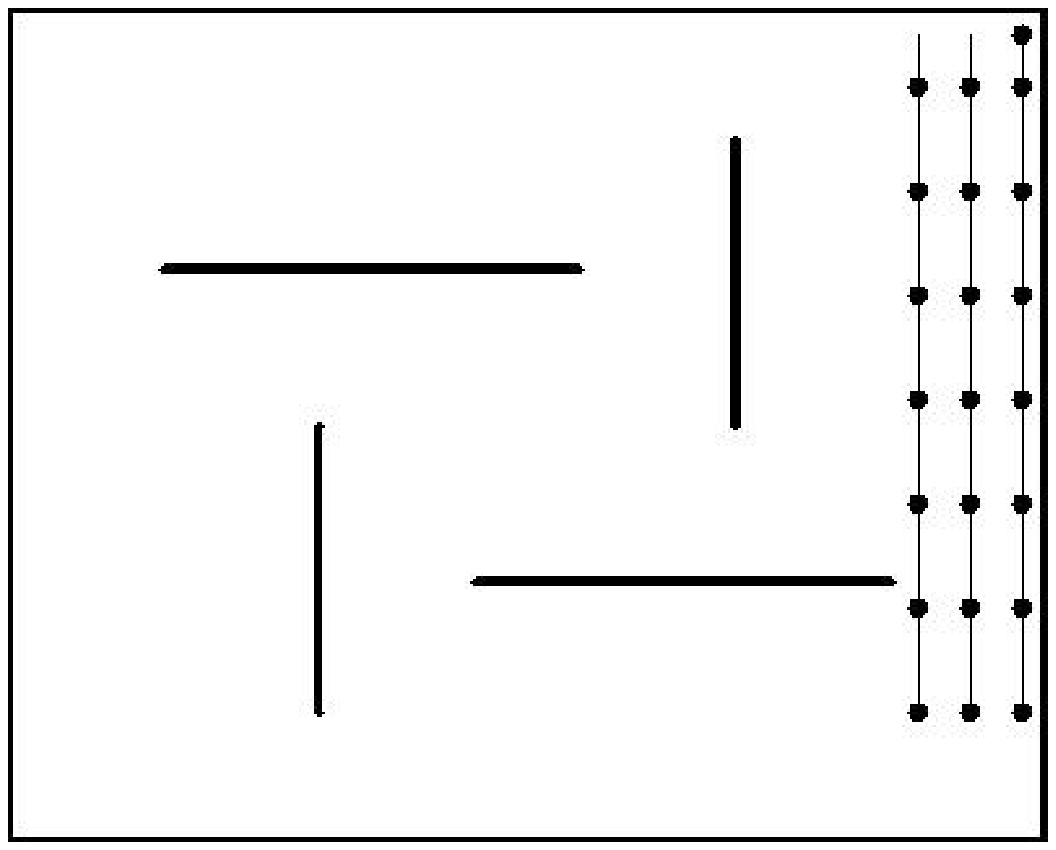}\hfill
\includegraphics[scale=.20]{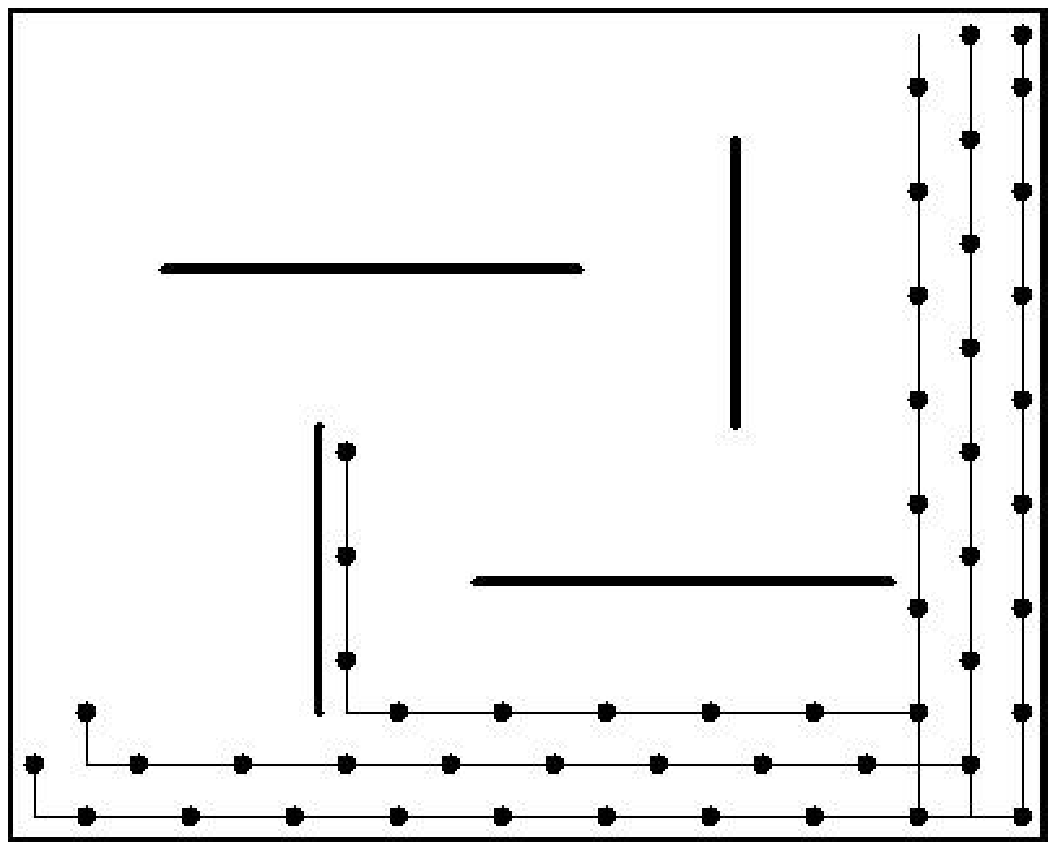}\hfill
\includegraphics[scale=.20]{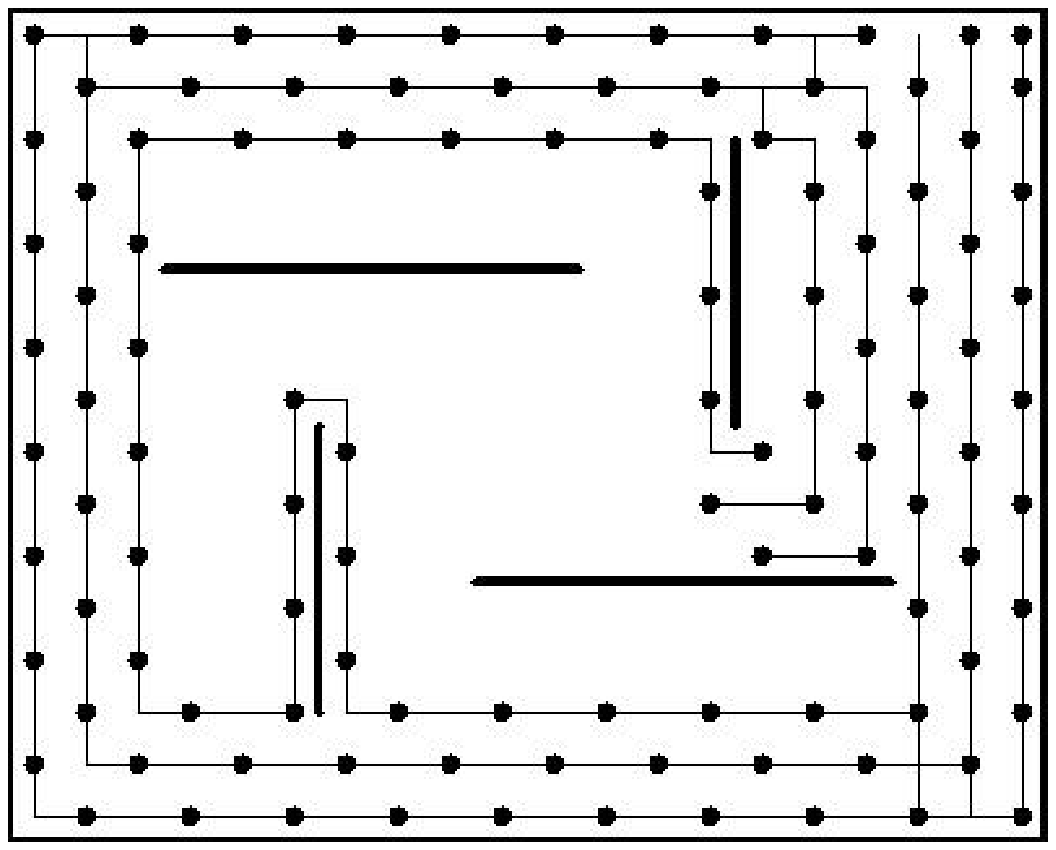}\hfill
\includegraphics[scale=.20]{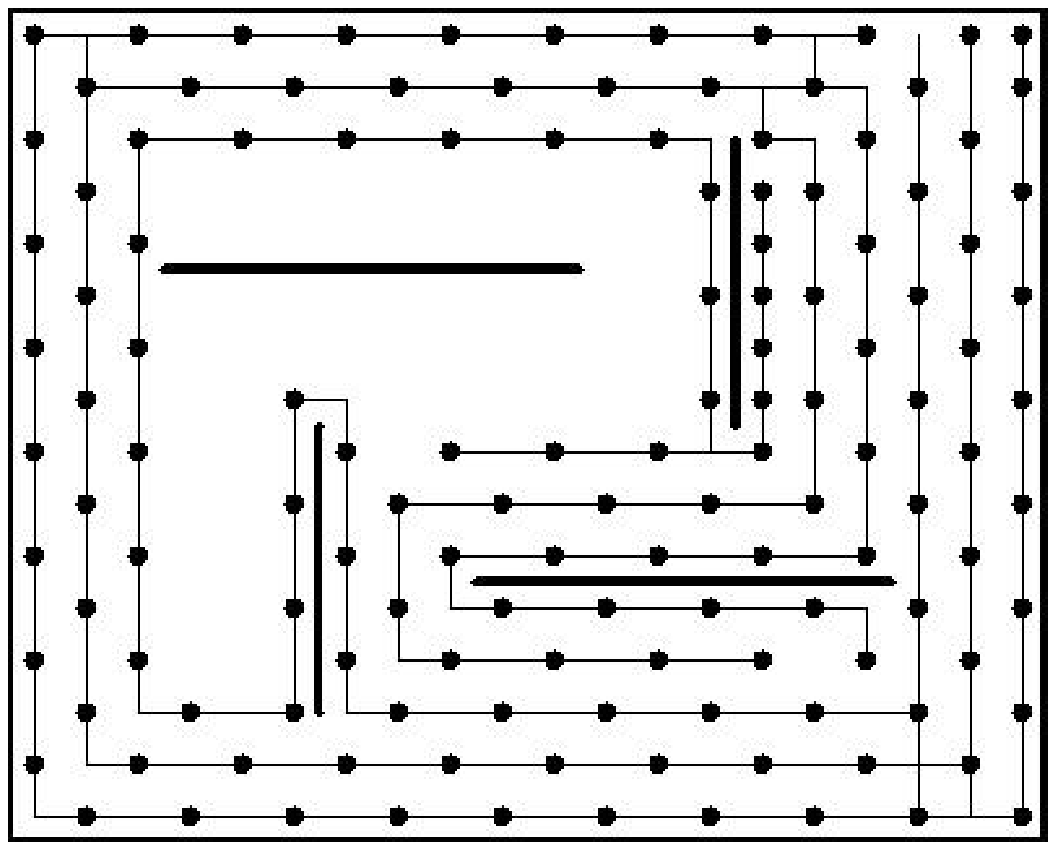}\hfill
\includegraphics[scale=.20]{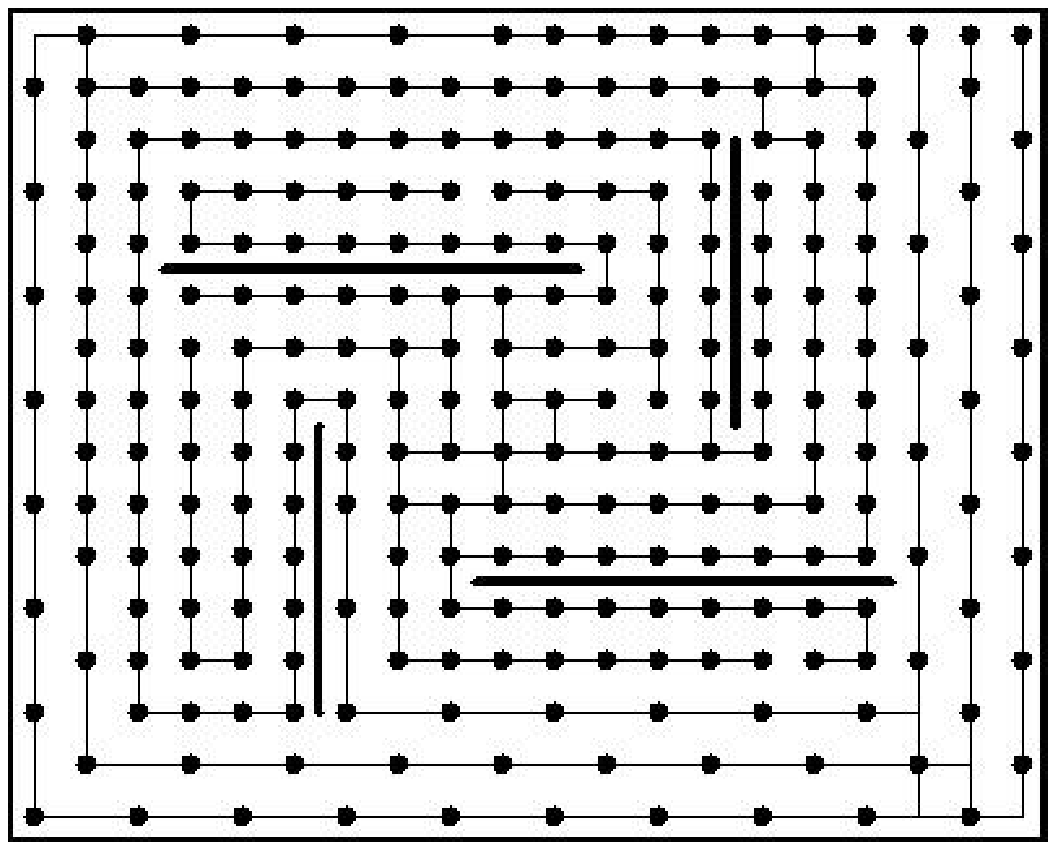}}
\label{fig:4walls}
\end{figure}

\begin{figure}[h!bt]
\centerline{
\includegraphics[scale=.20]{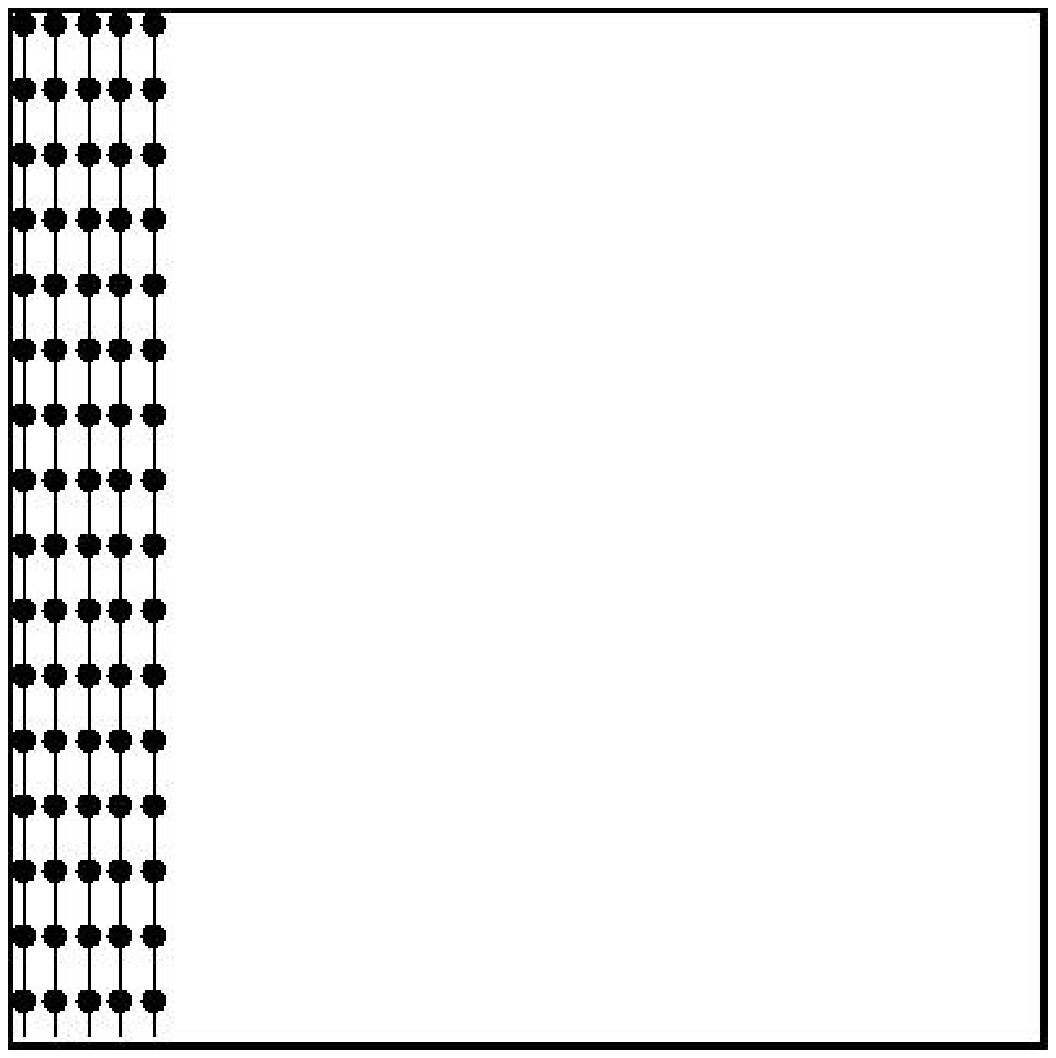}\hfill
\includegraphics[scale=.20]{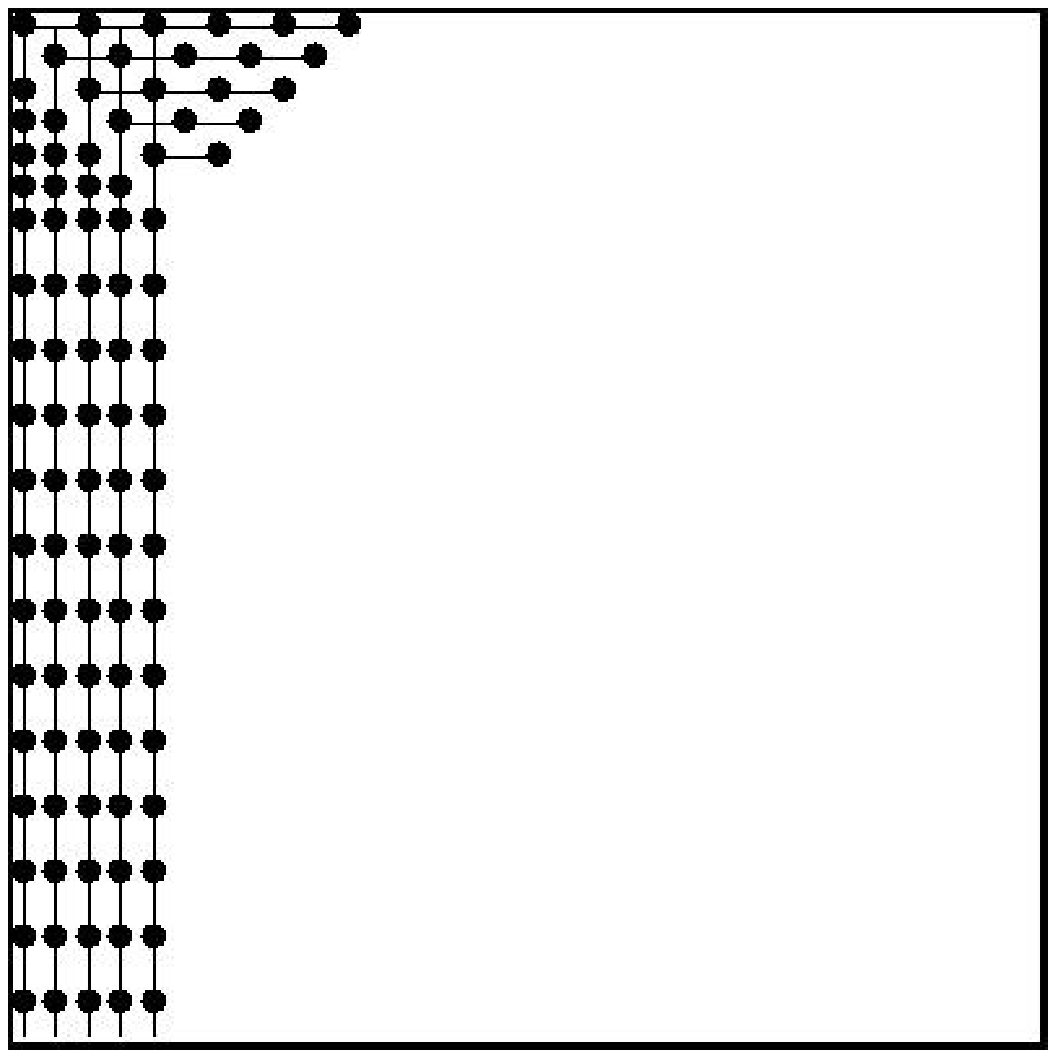}\hfill
\includegraphics[scale=.20]{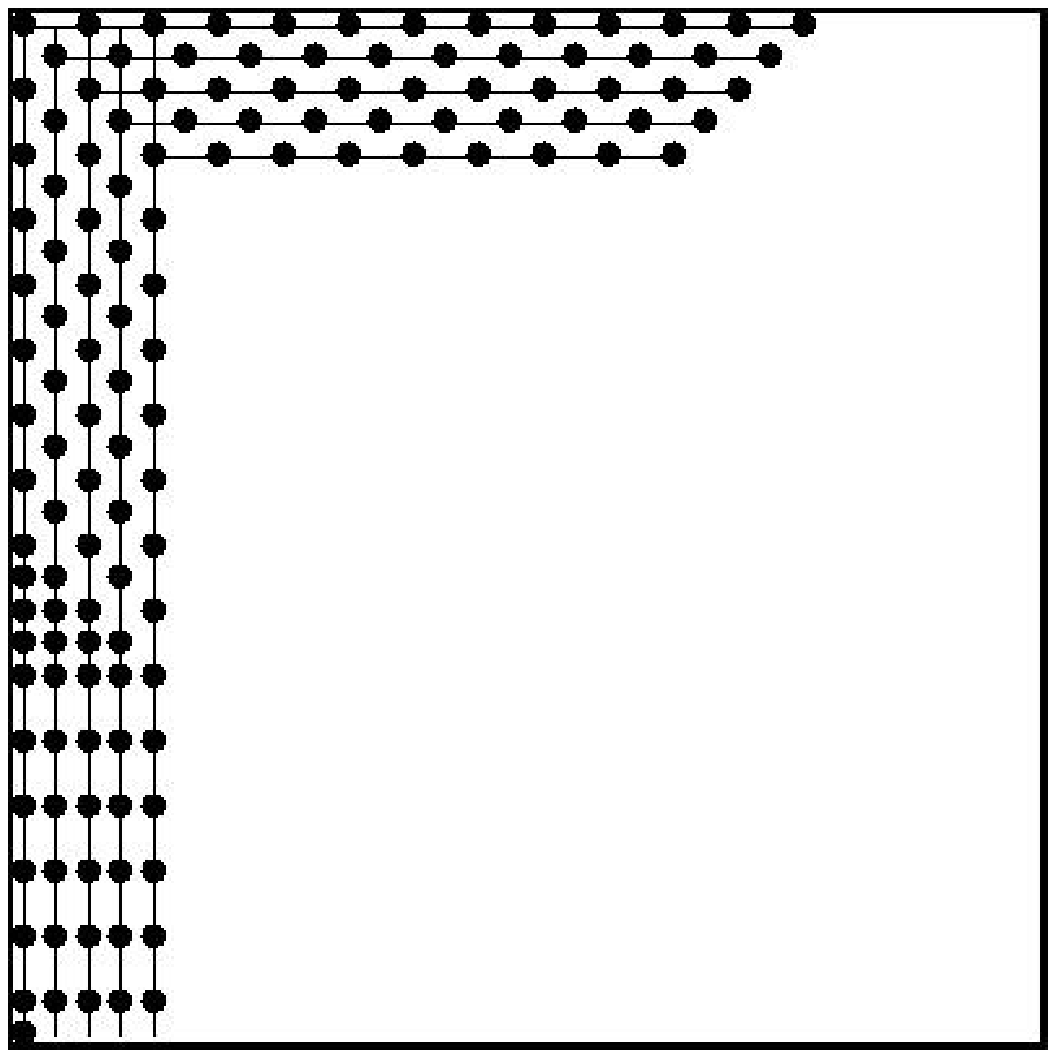}\hfill
\includegraphics[scale=.20]{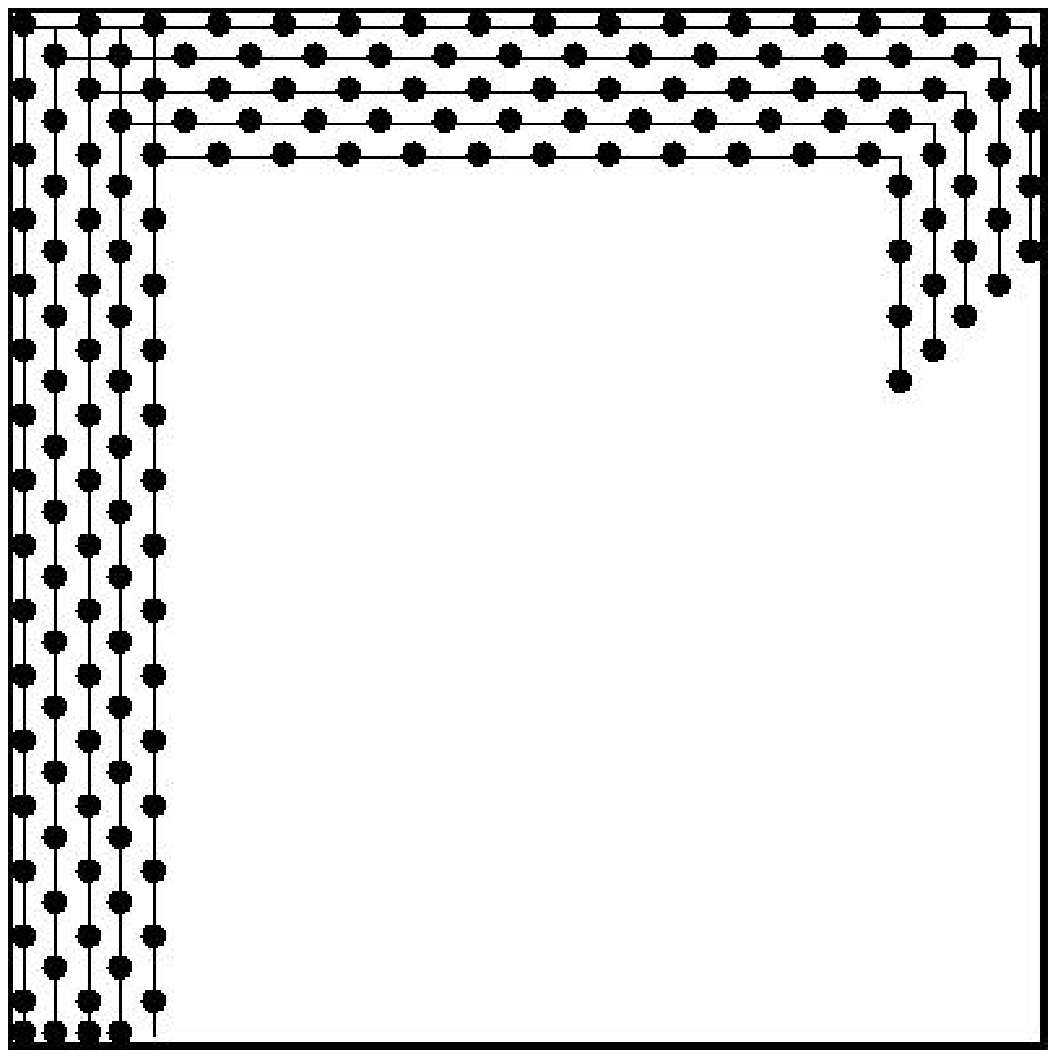}\hfill
\includegraphics[scale=.20]{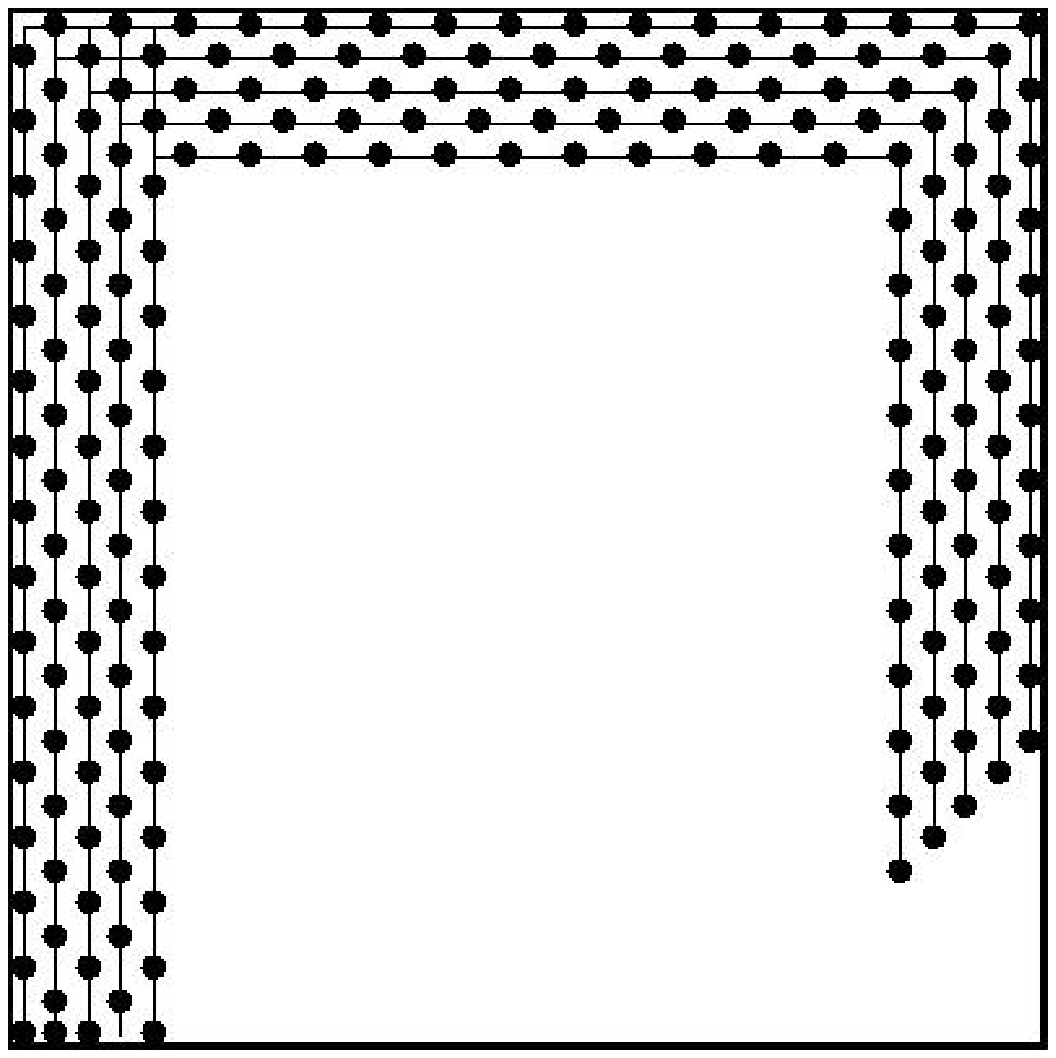}}
\caption{The LFLF algorithm: 
 There are six door pixels in the lower left corner. Splicing enables the flow to turn
a corner.}
\label{fig:square}
\end{figure}

\subsection{Correctness of LFLF}

We now show that LFLF always fills an environment.
During LFLF, we claim that the following invariant is maintained:
{\em The left side of a flow line consists of visited pixels and obstacles.}
This follows from the fact that, whenever a robot enters a pixel,
it turns as far left as it can (left-hand-on-the-wall).
The invariant implies the following lemma, which we use to
 show that the visited regions propagate
out from obstacle pixels:

\begin{lemma}
The union of boundary points separates the visited region from
obstacles and the right side of flows.
In other words, the unvisited region only touches a flow from its right.
\end{lemma}

Another useful property of  flow lines is the following:

\begin{lemma}
Flows do not self-intersect.
\end{lemma}

\begin{lemma}
Leadership passing does not change the total number of leader robots.
Thus, the number of leader robots at any time step $t$ is equal to
the number of active doors.
\end{lemma}

\begin{theorem}
The robots always fill environment $R$.
\end{theorem}

\begin{proof}
When a robot becomes inactive, it is a leader and its surrounding pixels
must all not be frontier, for otherwise $r$ would advance to such a pixel
and still be active. Therefore, if an active pixel $u$ is next to a
frontier pixel $v$, any robot advancing into $u$ will not become inactive
and some flow lines can advance into the empty region so it will be
filled.\qed
\end{proof}
In the next sections we analyze the makespan for this filling strategy.

\subsection{Competitive Analysis for Online Strategies}

For any strategy ${\cal S}$, let $n_{\cal S}(t)$ be the number of robots
that enter through the doors during timestep $t$;
we define $n_{\cal S}(0)=k$.

Note that the inherent delay of $2$ in robot movement means that
$n_{\cal S}(t)$
is a highly erratic function, so that in one timestep $k$ robots
might enter the doors, implying that in the next timestep
no robots enter the doors.
In LFLF
additional delays are caused by splicings, and therefore
we cannot even say that a robot enters an active door pixel every other
timestep.
To understand why, consider
Figure~\ref{fig:square}. When one flow line splices into a different
flow line, a robot may get delayed by one additional time step in
order to obtain the required separation between robots. If many
splicing are happening next to each other, this can cause an
accumulated delay of up to $k$ time units, which causes a ``wave'' of
packed robots that propagates back towards the door pixels. Thus, a
door pixel might temporarily hold off ejecting robots for many units
without the door pixel becoming inactive.
Whenever an active door pixel does not eject a robot we say there
is a {\em door-pixel delay\/}.

Fortunately, the following lemma bounds the number of splicings
(and also the total number of door-pixel delays).

\begin{lemma}
A pixel can only be used to initiate $O(1)$ splicings.
\end{lemma}

\begin{theorem}
The LFLF strategy is $O(\log (k+1))$-competitive.
\end{theorem}

\begin{proof}
Consider the times
$\{t_i\}_{i=0}^{\lceil\log (k+1)\rceil}$, 
where $t_i$ is the
latest timestep during which at least $k/2^i$ door pixels are active.
We call times $t_0, t_1, t_2, \ldots, t_{\lceil\log (k+1)\rceil}$
{\em significant events\/}.
Consider the time interval from
$t_i$ to $t_{i+1}$, when at least $k/2^{i+1}$ door pixels
are active.  Thus, the total number of robots leaving the door
plus door-pixel delays during
this interval is at least $(k/2^{i+1})(t_{i+1}-t_i)$.

Now consider an optimal strategy ${\cal S}^*$, and consider the cut
$\chi$ associated with the significant event $t_i$  of LFLF.
Cut $\chi$ is the boundary between the occupied and frontier
pixels and has size at most $k/2^i$.
A lower bound on the makespan of
any strategy is the area ``behind'' $\chi$ (i.e., the side of $\chi$
not containing the doors) divided by the length of $\chi$ plus
the amount of time required to send the flow from the door to reach
$\chi$.
Thus, the makespan, $\opt$, of ${\cal S}^*$ satisfies
$$\opt \geq \frac{{k}/ {2^{i+1}}(t_{i+1}-t_i)}{{k}/ {2^i}}= (t_{i+1}-t_i)/2
.$$
Summing over all significant events we obtain that the total
filling time of LFLF is $O(\opt\cdot\log (k+1))$.\qed
\end{proof}

If we disregard the issue of delays caused by pixels having to be fully
vacated before being re-entered, we get a whole class called
{\em sensible strategies},
defined by the following two conditions:

\begin{description}

\item[(1)]
$n_{\cal S}(t)$ is non-increasing in $t$; and
\item[(2)]
If $n_{\cal S}(t)>n_{\cal S}(t+1)$, then at time $t$ the number of
occupied pixels bordering on a frontier pixel is $n_{\cal S}(t+1)$.
\end{description}

The above proof also implies the following;
quite clearly, this result still holds if we extend the notion of
sensible strategies to allow for some delays, as
long as these only add a constant factor to the filling times.

\begin{theorem}
Any sensible strategy is $O(\log (\frac{k}{b}+1))$-competitive.
\end{theorem}

\begin{figure}[htbp]
\centerline{
\includegraphics[width=.4\textwidth]{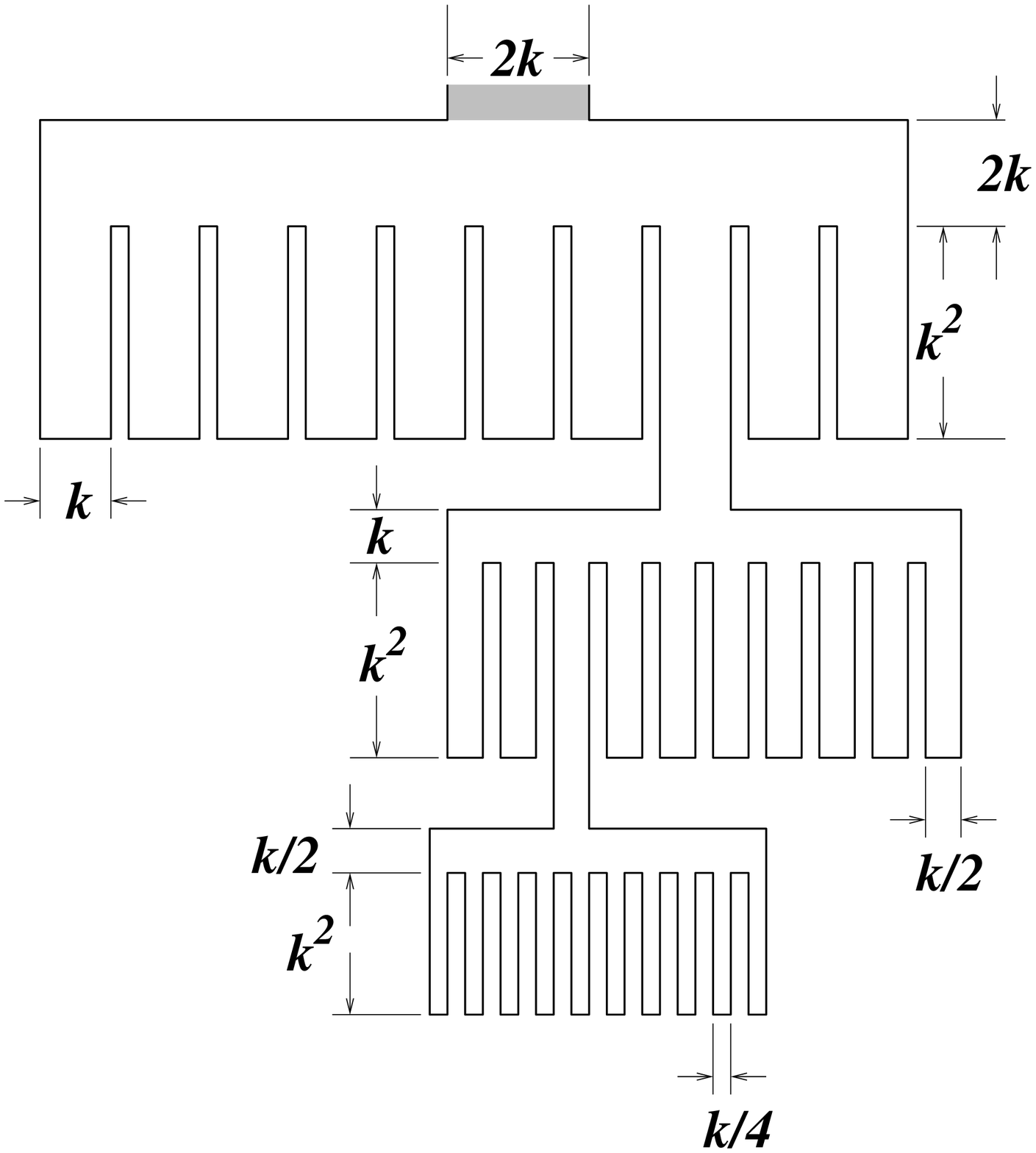}\hfill
\includegraphics[width=.45\textwidth]{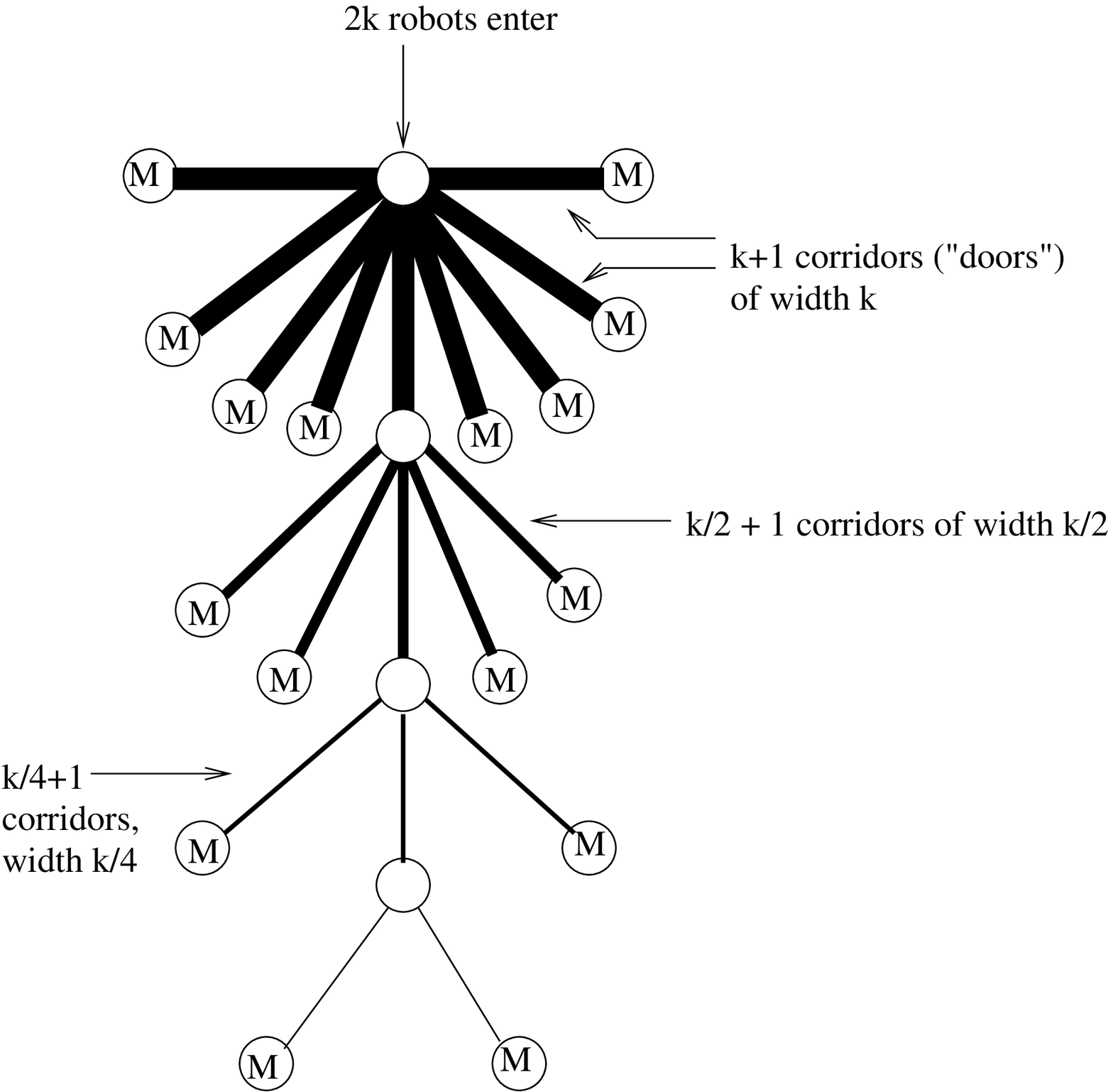}}
\caption{Left: An environment that is hard for both
depth-first  and breadth-first filling strategies.
This example shows that LFLF and related strategies are  $\Omega (\log (k+1))$
competitive.
Right: A scenario in which any simple-minded strategy
is $\Omega (\log (k+1))$ competitive.
}
\label{fig:log-comp}
\end{figure}

\subsection{Competitive Analysis of Simple-Minded Strategies}

Under certain natural assumptions that limit the power of strategies,
it is possible to give matching lower bounds on the competitiveness.
To give some intuition, consider first the class of examples shown in
Figure~\ref{fig:log-comp}(left), which
shows that the laminar flow algorithm may take as much as
$\Omega(\log (k+1))$ times $\opt$.

\begin{theorem}
The LFLF algorithm is $\Omega(\log (k+1))$-competitive.
\end{theorem}

This lower bound can be generalized to a much larger class of
strategies.  We say that strategy ${\cal S}$ is {\em simple-minded\/}
if the following condition holds: {\em There is a constant $C$, such
  that we can only tell the number $A'$ of pixels in a region
  $R'\subseteq R$ when a set $R''\subseteq R$ has been visited that
  has all pixels in $R'$ within $L_1$ distance~$C$.}

Considering simple-minded strategies is quite natural in the context
of robot swarms: If we assume we only ``know'' a region $R'$ after
each pixel has been seen by one of the robots, then the constant $C$
corresponds precisely to the sensor range.  The example of
Figure~\ref{fig:log-comp}(left) can be generalized as shown
schematically in Figure~\ref{fig:log-comp}(right), which forms the
basis for the proof of the following theorem:

\begin{theorem}
\label{th:lower.simple}
Any simple-minded strategy is $\Omega(\log (k+1))$-competitive.
\end{theorem}

A particular subclass of simple-minded strategies is one that
arises quite naturally in exploration problems:
We say that a strategy is {\em conservative} if any pixel
that has been ``discovered'', i.e., come within sensor range
of a robot, must stay within sensor range of some robot.
For a constant-size sensor range, it is easy to prove the following
statement, which has been claimed for a long time in a different context:

\begin{proposition}
\label{prop:lower.conserve}
All conservative strategies are simple-minded.
\end{proposition}

As the LFLF strategy is conservative by design, we conclude
that the lower bound of Theorem~\ref{th:lower.simple} applies to LFLF.

\subsubsection*{Acknowledgements}
We thank N.~Jovanovic and M.~Sztainberg for contributions to
this research.  This reseearch was partially supported by HRL Labs (a
DARPA subcontract), Honda Fundamental Research Labs, NASA Ames
Research (NAG2-1325), NSF (CCR-0098172, EIA-0112849, CCR-0208670), U.S.-Israel
Binational Science Foundation, and Sandia National Labs.

\end{document}